\documentclass[11pt,a4paper]{article}
\usepackage[hyperref]{acl2019}
\usepackage{times}
\usepackage{latexsym}
\usepackage{graphicx}
\usepackage{url}
\usepackage{amsfonts}
\usepackage{setspace}
\usepackage{amsmath}

\aclfinalcopy 
\def\aclpaperid{23} 

\title{\textit{``My Way of Telling a Story":} Persona based Grounded Story Generation}

\author{Shrimai Prabhumoye\thanks{ \hspace{0.45em}Both authors contributed equally to this work.} , Khyathi Raghavi Chandu\footnotemark[1]  ,  \textbf{Ruslan Salakhutdinov, Alan W Black}  \\
  Language Technologies Institute, Carnegie Mellon University \\
  Pittsburgh, PA, USA \\
  \texttt{ \{kchandu, sprabhum, rsalakhu,  awb\}@cs.cmu.edu} \\
  }

\date{}

\begin{document}
\maketitle
\begin{abstract}

Visual storytelling is the task of generating stories based on a sequence of images.
Inspired by the recent works in neural generation focusing on controlling the \textit{form} of text, this paper explores the idea of generating these stories in different personas. However, one of the main challenges of performing this task is the lack of a dataset of visual stories in different personas. Having said that, there are independent datasets for both visual storytelling and annotated sentences for various persona. In this paper we describe an approach to overcome this by getting labelled persona data from a different task and leveraging those annotations to perform persona based story generation.
We inspect various ways of incorporating personality in both the encoder and the decoder representations to steer the generation in the target direction. To this end, we propose five models which are incremental extensions to the baseline model to perform the task at hand.
In our experiments we use five different personas to guide the generation process.
We find that the models based on our hypotheses perform better at capturing words while generating stories in the target persona.

\end{abstract}

\section{Introduction}

Storytelling through pictures has been dated back to prehistoric times -- around 30,000 years ago, paintings of herds of animals like bisons, rhinos and gazelles were made in a cave in Southern France.
However, these were not merely paintings, they were stories about the heroic adventures of humans.
Since then visual storytelling has evolved from paintings to photography to motion pictures to video games.
With respect to its timeline, neural generative storytelling has gained traction only recently.
Recent research has focused on challenges in generating longer documents \cite{wiseman2017challenges, lau2016empirical} as well as on predicting the next events in the story \cite{martin2018event}.
Contemporary research has focused on using deep generative models to capture high-level plots and structures in stories \cite{fan2018hierarchical}. Recent years have also seen some work hinging on the event structures and scripts \cite{mostafazadeh2016caters, rishes2013generating, peng2018towards}. Generating an appropriate ending of a story was also studied by \citet{guan2018story} and \citet{sharma2018tackling}. Research on generating stories from a sequence of images is anew \cite{peng2018towards, lukin2018pipeline, kim2018glac, hsu2018using, gonzalez2018contextualize}.

\citet{cavazza2009emotional} have stressed the importance of expressing emotions in the believability of the automated storytelling system.
Adapting a personality trait hence becomes crucial to capture and maintain interest of the audience.
Associating the narrative to a personality instigates a sense of empathy and relatedness.
Although there has been research in generating persona based dialog responses and generating stylistic sentences \cite{shuster2018engaging, fu2018style, prabhumoye2018style, shen2017style}, generating persona based stories with different personality types narrating them has been unexplored.
In this paper, we focus on generating a story from a sequence of images as if the agent belongs to a particular personality type. In specific, we choose to perform experimentations on visual story telling \cite{huang2016visual}.

This paper introduces a novel approach to generating visual stories in five different personality types. A key challenge to this end is the lack of large scale persona annotated stories. We address this by transferring knowledge from annotated data in dialog domain to the storytelling domain.
We base our visual story generator model on \citet{kim2018glac} and propose multiple techniques to induce the personalities in the latent representations of both the encoder and the decoder.
The goal of our work is to learn the mapping between the latent representations of the images and the tokens of the story such that we encourage our generative model to generate tokens of a particular personality.
We evaluate our generative models using the automatic metric of ROUGE \cite{lin2004rouge} which takes into account the sentence level similarity in structure and thus roughly evaluates the matching of content. We acknowledge that there is a drop in this metric since our model is not trying to optimize generation alone but also adapt personality from a different dataset.

We also evaluate the success of generating the story in the target personality type using automatic and qualitative analysis. The automatic metrics comprise of the classification accuracies rooted from the annotated data. We observe that one of the proposed models (LEPC, described in Section \ref{sec:models} performs slightly better at classification accuracies for most of the personas while retaining similar ROUGE scores.

The main contribution of this paper is showing simple yet effective approaches to narrative visual stories in different personality types.
The paper also displays an effective way of using annotated data in the dialog domain to guide the generative models to a specified target personality.

\section{Related Work}


\paragraph{Visual Story Telling: }
Last decade witnessed enormous interest in research at the intersection of multiple modalities, especially vision and language.
Mature efforts in image captioning \cite{hossain2019comprehensive} paved way into more advanced tasks like visual question answering \cite{wu2017visual} and visual dialog  \cite{das2017visual} , \cite{mostafazadeh2017image}.
As an obvious next step from single shot image captioning lies the task of describing a sequence of images which are related to one another to form a story like narrative.
This task was introduced as visual story telling by \citet{huang2016visual}, differentiating descriptions of images in isolation (image captions) and stories in sequences.
The baseline model that we are leveraging to generate personality conditioned story generation is based on the model proposed by \citet{kim2018glac} for the visual story telling challenge.
Another simple yet effective technique is late fusion model by \citet{smilevski2018stories}.
In addition to static images, \citet{gella2018dataset} have also collected a dataset of describing stories from videos uploaded on social media. \citet{chandustoryboarding} recently introduced a dataset for generating textual cooking recipes from a sequence of images and proposed two models to incorporate structure in procedural text generation from images.

\paragraph{Style Transfer: } One line of research that is closely related to our task is style transfer in text.
Recently generative models have gained popularity in attempting to solve style transfer in text with non-parallel data \cite{hu2017toward, shen2017style, li2018delete}.
Some of this work has also focused on transferring author attributes \cite{prabhumoye2018style}, transferring multiple attributes \cite{lample2018multipleattribute, logeswaran2018content} and collecting parallel dataset for formality \cite{rao2018dear}.
Although our work can be viewed as another facet of style transfer, we have strong grounding of the stories in the sequence of images.

\paragraph{Persona Based Dialog: }
Persona based generation of responses has been studied by NLP community in dialog domain.
\cite{li2016persona} encoded personas of individuals in contextualized embeddings that capture the background information and style to maintain consistency in the responses given.
The embeddings for the speaker information are learnt jointly with the word embeddings.
Following this work, \cite{zhou2018emotional} proposed Emotional Chatting Machine that generates responses in an emotional tone in addition to conditioning the content.
The key difference between former and latter work is that the latter captures dynamic change in emotion as the conversation proceeds, while the user persona remains the same in the former case.
\cite{zhang2018personalizing} release a huge dataset of conversations conditioned on the persona of the two people interacting.
This work shows that conditioning on the profile information improves the dialogues which is measured by next utterance prediction.
In these works, the gold value of the target response was known.
For our work, we do not have gold values of stories in different personas. Hence we leverage annotated data from a different task and transfer that knowledge to steer our generation process.

\paragraph{Multimodal domain: }
With the interplay between visual and textual modalities, an obvious downstream application for persona based text generation is image captioning.
\citet{chandrasekaran2018punny} worked on generating witty captions for images by both retrieving and generating with an encoder-decoder architecture.
This work used external resources to gather a list of words that are related to puns from web which the decoder attempts to generate conditioned on phonological similarity.
\citet{wang2015can} studied the statistical correlation of words associated with specific memes.
These ideas have also recently penetrated into visual dialog setting. \citet{shuster2018engaging} have collected a grounded conversational dataset with 202k dialogs where humans are asked to portray a personality in the collection process.
They have also set up various baselines with different techniques to fuse the modalities including multimodal sum combiner and multimodal attention combiner. We use this dataset to learn personas which are adapted to our storytelling model.

\section{Models}
\label{sec:models}

We have a dataset of visual stories $\boldsymbol{S} = \{\boldsymbol{S_1}, \ldots, \boldsymbol{S_n} \}$.
Each story $\boldsymbol{S_i}$ is a set of sequence of five images and the corresponding text of the story $\boldsymbol{S_i} = \{(\boldsymbol{I}_i^{(1)}, \boldsymbol{x}_i^{(1)}), \ldots, (\boldsymbol{I}_i^{(5)}, \boldsymbol{x}_i^{(5)})\}$.
Our task is to generate the story based on not only the sequence of the images but also closely following the narrative style of a personality type.
We have five personality types (described in Section \ref{sec:data}) $\boldsymbol{P} = \{\boldsymbol{p}_1, \ldots, \boldsymbol{p}_5\}$ and each story is assigned one of these five personalities as their target persona.
Here, each $\boldsymbol{p}_i$ represents the one-hot encoding of the target personality for story i.e $\boldsymbol{p}_1 = [1, 0, 0, 0, 0]$ and so on till $\boldsymbol{p}_5 = [0, 0, 0, 0, 1]$.
Hence, we create a dataset such that for each story, we also have a specified target personality type $\boldsymbol{S}_i = \{(\boldsymbol{I}_i^{(1)}, \boldsymbol{x}_i^{(1)}), \ldots, (\boldsymbol{I}_i^{(5)}, \boldsymbol{x}_i^{(5)}); \boldsymbol{p}_i\}$.
The inputs to our models are the sequence of images and the target personality type.
We build generative models such that they are able to generate stories in the specified target personality type from the images. In this section, we first briefly describe classifiers that are trained discriminatively to identify each of the personalities and then move on to the story generation models that make use of these classifiers.

Here is an overview of the differences in the six models that we describe next.
\begin{enumerate}
    \item The baseline model (Glocal) is a sequence to sequence model with global and local contexts for generating story sentence corresponding to each image.
    \item The Multitask Personality Prediction (MPP) model is equipped with predicting the personality in addition to generating the sentences of the story. This model also incorporates binary encoding of personality.
    \item The Latent Encoding of Personality in Context (LEPC) model incorporates an embedding of the personality as opposed to binary encoding.
    \item The Latent Encoding of Personality in Decoder (LEPD) model augments personality embedding at each step in the decoder, where each step generates a token.
    \item  Stripped  Encoding  of  Personality  in  Context  (SEPC) is similar to LEPC but encodes personality embedding after stripping the mean of the story representation.
    \item Stripped  Encoding  of  Personality  in  Decoder  (SEPD) is similar to LEPD but encodes personality embedding after stripping the mean of the story representation. This is similar to the intuition behind SEPC.
\end{enumerate}

\subsection{Classification}
\label{sec:classification}
We use convolutional neural network (CNN) architecture to train our classifiers.
We train five separate binary classifiers for each of the personality types.
The classifiers are trained to predict whether a sentence belongs to a particular personality or not.
We train the classifiers in a supervised manner.
We need labeled data to train each of the classifiers.
Each sample of text $\boldsymbol{x}$ in the respective datasets of each of the five personality types has a label in the set $\{0, 1\}$.
Let $\boldsymbol{\theta}_{\text{C}}^{\boldsymbol{\boldsymbol{p}_j}}$ denote the parameters of the classifier for personality $\boldsymbol{p}_j$ where $j \in \{1, \ldots, 5\}$.
Each classifier is trained with the following objective:
\begin{equation}
\boldsymbol{\mathcal{{L}}} (\boldsymbol{\theta}_{\text{C}}^{\boldsymbol{p}_j}) = \mathbb{E}_{\boldsymbol{X}} [\log q_{C} (\boldsymbol{p}_j |\boldsymbol{x})]
\label{eq:class}
\end{equation}
We use cross entropy loss to calculate $\boldsymbol{\mathcal{{L}}}_{\text{C}}^{\boldsymbol{\boldsymbol{p}_j}}$ for each of the five classifiers.
The classifiers accept continuous representations of tokens as input.

\subsection{Story Generation}

We present five extensions to incorporate personality based features in the generation of stories.

\paragraph{(1) Baseline model (Glocal): }
We first describe the baseline model that is used for visual story telling.
This is based on the model \cite{kim2018glac} that attained better scores on human evaluation metrics.
It follows an encoder-decoder framework translating a sequence of images into a story.
From here on, we refer to this model as \textit{glocal} through the rest of the paper owing to the global and local features in the generation of story sequence at each step (described in this section).

The image features for each of the steps are extracted with a ResNet-152 \cite{he2016deep} post resizing to 224 X 224.
The features are taken from the penultimate layer of this pretrained model and the gradients are not propagated through this layer during optimization.
These features are passed through a fully connected layer to obtain the final image features.
In order to obtain an overall context of the story, the sequence of the image features are passed through a Bi-LSTM.
This represents the global context of the story.
For each step in the generation of the story, the local context corresponding to the specificity of that particular image is obtained by augmenting the image features (local context) to the context features from the Bi-LSTM (global context).
These \textit{glocal features} are used to decode the story sentence at each step. This concludes the encoder part of the story.
The decoder of each step in the story also uses an LSTM which takes the same glocal feature for that particular step at each time step.
Hence there are 5 glocal features feeding into each time step in the decoder.

For simplicity in understanding, we use the following notations throughout model descriptions to represent mathematical formulation of the generation models.
Subscript \textit{k} indicates the \textit{$k^{th}$} step or sentence in a story.
Subscript \textit{i} indicates the \textit{$i^{th}$} story example.
The story encoder is represented as \textit{Encoder} which comprises of the features extracted from the penultimate layer of ResNet-152  concatenated with the global context features from the Bi-LSTM.
The entirety of this representation in encoder and the glocal features obtained is represented using $\boldsymbol{z}_k$ for the \textit{$k_{th}$} step or sentence in the story.

\begin{equation}
    \boldsymbol{z}_k = \textit{Encoder} (\boldsymbol{I}_k)
\end{equation}

Now, the generation of a sentence in the story is represented as follows:

\begin{equation}
    \boldsymbol{\hat{x}}_k \sim \prod_t Pr(\boldsymbol{\hat{x}}_k^t | \boldsymbol{\hat{x}}_k^{<t}, \boldsymbol{z}_k)
\end{equation}

The generated sentence $\boldsymbol{\hat{x}}_k$ is obtained from each of the output words $\boldsymbol{\hat{x}}_k^t$ which is generated by conditioning on all of the prior words $\boldsymbol{\hat{x}}_k^{<t}$ and the glocal feature obtained as $\boldsymbol{z}_k$.

\paragraph{Personality based Generation: } In the rest of the section, we are going to describe the incremental extensions to the baseline to adapt the model to perform persona based story generation.

\paragraph{(2) Multitask Personality Prediction (MPP):} The intuition behind the hypothesis here is to provide the personality information to the model and also enable it to predict the personality along with the generation of the story.
The obvious extension to provide personality information is to incorporate the one-hot encoding $\boldsymbol{p}_i \in \boldsymbol{P}$ of the five personas in the context before the decoder.
The visual story telling data is split into five predetermined personalities as described in Section \ref{sec:data}. For each story, the corresponding personality is encoded in a one hot representation and is augmented to the glocal context features. These features are then given to the decoder to produce each step in the story.
The model is enabled to perform two tasks: the primary task is to generate the story and the secondary task is to predict the personality of the story. The classifiers described in Section \ref{sec:classification} are used to perform personality prediction. Formally, the generation process is represented by:

\begin{equation}
    \boldsymbol{\hat{x}}_k \sim \prod_t Pr(\boldsymbol{\hat{x}}_k^t | \boldsymbol{\hat{x}}_k^{<t}, \boldsymbol{z}_k, \boldsymbol{p}_i)
\end{equation}

Here, we condition the generation of each word on the glocal context features $\boldsymbol{z}_k$, binary encoding of the personality $\boldsymbol{p}_i$ and the words generated till that point.

 The cross entropy loss for generation is $\mathbf{\boldsymbol{\mathcal{L}_{g}}}$ and the loss for the prediction of each of the personalities is $\boldsymbol{L}_{\text{C}}^{\boldsymbol{\boldsymbol{p}_j}}$ given by Eq \ref{eq:class}. The overall loss optimized for this model is:

\begin{align*}
\boldsymbol{\mathcal{L}_{total}} = \alpha \cdot \boldsymbol{\mathcal{L}_{g}} + \frac{(1 - \alpha)}{5} \cdot \sum_{j=1}^{5} \boldsymbol{\mathcal{L}}_{\text{C}}^{\boldsymbol{\boldsymbol{p}_j}}
\end{align*}

The overall model is optimized on this total loss.
We use cross entropy loss for each of the individual losses.
We give a higher weight $\alpha$ to the story generation and equally distribute the remaining $(1 - \alpha)$ among each of the 5 personalities.

\paragraph{(3)  Latent Encoding of Personality in Context (LEPC): } This model is an incremental improvement over MPP model.
The key difference is the incorporation of personality as an embedding that captures more centralized traits in the words belonging to that particular personality.
For each of the five personality types,
we have a latent representation of the personality ($\boldsymbol{\mathcal{P}}$), as opposed to the binary encoding in MPP model. Similar to the earlier setting, this average personality feature vector is concatenated with the glocal context vector
The generation step is formally represented as:

\begin{equation}
    \boldsymbol{\hat{x}}_k \sim \prod_t Pr(\boldsymbol{\hat{x}}_k^t | \boldsymbol{\hat{x}}_k^{<t}, [\boldsymbol{z}_k; \boldsymbol{\mathcal{P}}], \boldsymbol{p}_i)
\end{equation}

This means that $\boldsymbol{z}_k$ is concatenated with $\boldsymbol{\mathcal{P}}$ to give personality informed representation; and the generation of each word is conditioned on these concatenated features $\boldsymbol{z}_k$, binary encoding of the personality $\boldsymbol{p}_i$ and the words generated so far.

\paragraph{(4) Latent Encoding of Personality in Decoder (LEPD): } Instead of augmenting the personality traits to the context as done in LEPC model, they could be explicitly used in each step of decoding.
The latent representation of the personality ($\boldsymbol{\mathcal{P}}$) is concatenated with the word embedding for each time step in the decoder.

\begin{equation}
    \boldsymbol{\hat{x}}_k \sim \prod_t Pr(\boldsymbol{\hat{x}}_k^t | [\boldsymbol{\hat{x}}_k^{<t}; \boldsymbol{\mathcal{P}}], \boldsymbol{z}_k, \boldsymbol{p}_i)
\end{equation}

The generation of each of the words is conditioned on the words generated so far that are already concatenated with the average vector for the corresponding personality, and the glocal features along with the binary encoding of the personality.

\paragraph{(5) Stripped Encoding of Personality in Context (SEPC): } In order to orient the generation more towards the personality, we need to go beyond simple augmentation of personality. Deriving motivation from neural storytelling\footnote{\url{https://github.com/ryankiros/neural-storyteller}}, we use a similar approach to subtract central characteristics of words in a story and add the characteristics of the personality.
Along the same lines of calculating an average representation for each of the personalities, we also obtain an average representation of the story $\boldsymbol{\mathcal{S}}$.
This average representation $\boldsymbol{\mathcal{S}}$ intuitively captures the style of the story.
Essentially, the story style is being stripped off the context and personality style is incorporated.
The modified glocal feature that is given to the decoder is obtained as
$\boldsymbol{m} = \boldsymbol{z}_k - \boldsymbol{\mathcal{S}} + \boldsymbol{\mathcal{P}} $.
The generation process is now conditioned on $\boldsymbol{m}$ instead of $\boldsymbol{z}_k$. 
Hence, the generation of each word in decoding is conditioned on the words generated so far ($\boldsymbol{\hat{x}}_k^{<t}$), the binary encoding of the personality ($\boldsymbol{p}_i$) and the modified representation of the context features ($\boldsymbol{m}$).

\begin{equation}
    \boldsymbol{\hat{x}}_k \sim \prod_t Pr(\boldsymbol{\hat{x}}_k^t | \boldsymbol{\hat{x}}_k^{<t}, \boldsymbol{m}, \boldsymbol{p}_i)
\end{equation}

Here, note that the context features obtained thus far are from the visual data and performing this operation is attempting to associate the visual data with the central textual representations of the personalities and the stories.

\paragraph{(6) Stripped Encoding of Personality in Decoder (SEPD): } This model is similar to SEPC with the modification of performing the stripping at each word embedding in the decoder as opposed to the context level stripping. The time steps to strip features is at the sentence level in SEPC and is at word level in SEPD model. The LSTM based decoder decodes one word at a time. At each of these time steps, the word embedding feature $\boldsymbol{\mathcal{E}}$ is modified as $\boldsymbol{e}_k = \boldsymbol{\mathcal{E}} - \boldsymbol{\mathcal{S}} + \boldsymbol{\mathcal{P}} $.
This modification is performed in each step of the decoding process.
These modified features are used to generate each sentence in the full story. The model is trained to generate a sentence in the story as described below:

\begin{equation}
    \boldsymbol{\hat{x}}_k \sim \prod_t Pr(\boldsymbol{\hat{x}}_k^t | \boldsymbol{e}_k^{<t}, \boldsymbol{z}_k, \boldsymbol{p}_i)
\end{equation}

The generation of each word is conditioned on the modified word embeddings using the aforementioned transformation ($\boldsymbol{e}_k^{<t}$), the binary encodings of the personalities ($\boldsymbol{p}_i$) and the glocal context features.

\section{Datasets}
\label{sec:data}

Coalescing the segments of personality and sequential generation together, our task is to generate a grounded sequential story from the view of a personality. To bring this to action, we describe the two sources of data we use to generate personality based stories in this section. The first source of data is focussed on generic story generation from a sequence of images and the second source of data includes annotations for personality types for sentences. We tailor a composition of these two sources to obtain a dataset for personality based visual storytelling. Here, we note that the techniques described above can be applied for unimodal story generation as well.

\paragraph{Visual Story Telling: }

Visual Storytelling is the task of generating stories from a sequence of images. A dataset for this grounded sequential generation problem was collected by  \citet{huang2016visual} and an effort for a shared task \footnote{\url{http://visionandlanguage.net/workshop2018/index.html\#challenge}} was led in 2018. The dataset includes 40,155 training sequences of stories. It comprises of a sequence of images, descriptions of images in isolation and stories of images in sequences. We randomly divide the dataset into 5 segments (comprising of 8031 stories each) and each segment is associated with a personality.

\paragraph{Personality Dialog: }\citet{shuster2018engaging} have provided a dataset of 401k dialog utterances, each of which belong to one of 215 different personalities.
The dataset was collected through image grounded human-human conversations.
Humans were asked to play the role of a given
personality.
This makes this dataset very pertinent for our task as it was collected through engaging image chat between two humans enacting their personalities.

For our task, we wanted to choose a set of five distinct personality types.
Let the set of utterances that belong to each personality type be $U_p = \{u_p^{1}, \ldots, u_p^{n}\}$ where $p \in \{1, \ldots, 215\}$.
We first calculate the pooled BERT representation \cite{devlin2018bert} of each of the utterances.
To get the representation of the personality $\boldsymbol{\mathcal{P}}$, we simply average the BERT representations of all the utterances that belong to that personality.
The representation of each personality is given by:
\begin{equation}
  \boldsymbol{\mathcal{P}}_p = \frac{\Sigma_{k=1}^{n} BERT(u_{p}^{k})}{n}\\
  \label{eq:perona_rep}
\end{equation}
This representation is calculated only on the train set of \cite{shuster2018engaging}.

Since our goal is to pick five most distinct personality types, we have the daunting task of filtering the 215 personality types to 5.
To make our task easier we want to group similar personalities together.
Hence, we use K-Means Clustering to cluster the representations of the personalities into 40 clusters \footnote{We do not perform exhaustive search on the number of clusters. We tried $k$ values of 5, 20 and 40 and selected 40 as the ideal value based on manual inspection of the clusters.}.
We get well formed and meaningful clusters which look like [Impersonal, Aloof (Detached, Distant), Apathetic (Uncaring, Disinterested), Blunt, Cold, Stiff]; [Practical, Rational, Realistic, Businesslike]; [Empathetic, Sympathetic, Emotional]; [Calm, Gentle, Peaceful, Relaxed, Mellow (Soothing, Sweet)] etc.
We then build a classifier using the technique described in Section \ref{sec:classification} to classify the utterances to belong to one of the 40 clusters.
We pick the top five clusters that give the highest accuracy for the 40-way classification.

 The five personality clusters selected are:
\begin{itemize}
    \item Cluster 1 \textbf{(C1)}: Arrogant, Conceited, Egocentric, Lazy, Money-minded, Narcissistic, Pompous and Resentful
    \item Cluster 2 \textbf{(C2)}: Skeptical and Paranoid
    \item Cluster 3 \textbf{(C3)}: Energetic, Enthusiastic, Exciting, Happy, Vivacious, Excitable
    \item Cluster 4 \textbf{(C4)}: Bland and Uncreative
    \item Cluster 5 \textbf{(C5)}: Patriotic
\end{itemize}

We build five separate classifiers, one for each personality cluster. Note that these clusters are also associated with personalities and hence are later referred as P followed by the cluster id in the following sections.
To build the five binary classifiers, we create label balanced datasets for each cluster i.e we randomly  select as many negative samples from the remaining 4 clusters as there are positive samples in that cluster.
We use the train, dev and test split as is from \cite{shuster2018engaging}.
The dataset statistics for each of the five clusters is provided in Table \ref{tab:stat}.
\begin{table}[h]
\centering
\begin{tabular}{l | r | r | r }
\hline
Cluster Type & Train & Dev & Test \\
\hline
Cluster 1 & 26538 & 1132 & 2294 \\
Cluster 2 & 6614 & 266 & 608 \\
Cluster 3 & 19784 & 898 & 1646 \\
Cluster 4 & 6646 & 266 & 576 \\
Cluster 5 & 3262 & 138 & 314 \\
\hline
\end{tabular}
\caption{Statistics of data belonging to each of the persona clusters}
\label{tab:stat}
\end{table}

Note that all the datasets have a balanced distribution of labels $0$ and $1$.
For our experiments it does not matter that distribution of the number of samples is different because we build separate classifiers for each of the cluster and their output is treated as independent from one another.

As seen in Table \ref{tab:class_stat}, all the classifiers attain good accuracies and F-scores on the test set.
\begin{table}[h]
\centering
\begin{tabular}{l | r | r | r | r | r }
\hline
 & C1 & C2 & C3 & C4 & C5 \\
\hline
Acc. & 79.12 & 81.09 & 83.17 & 77.95 & 84.08 \\
F1 & 0.79 & 0.81 & 0.83 & 0.78 & 0.84 \\
\hline
\end{tabular}
\caption{Performance of classifiers for each of the persona clusters }
\label{tab:class_stat}
\end{table}

We finally calculate the representation $\boldsymbol{\mathcal{P}}$ for each of the five clusters and the representation $\boldsymbol{\mathcal{S}}$ of stories using equation \ref{eq:perona_rep}.
Note that $\boldsymbol{\mathcal{S}}$ is calculated over the visual story tellind dataset.
These representations are used by our generative models \textbf{LEPC}, \textbf{LEPD}, \textbf{SEPC}, and \textbf{SEPD}.

\begin{figure*}[h!]
\centering
\includegraphics[width=\linewidth]{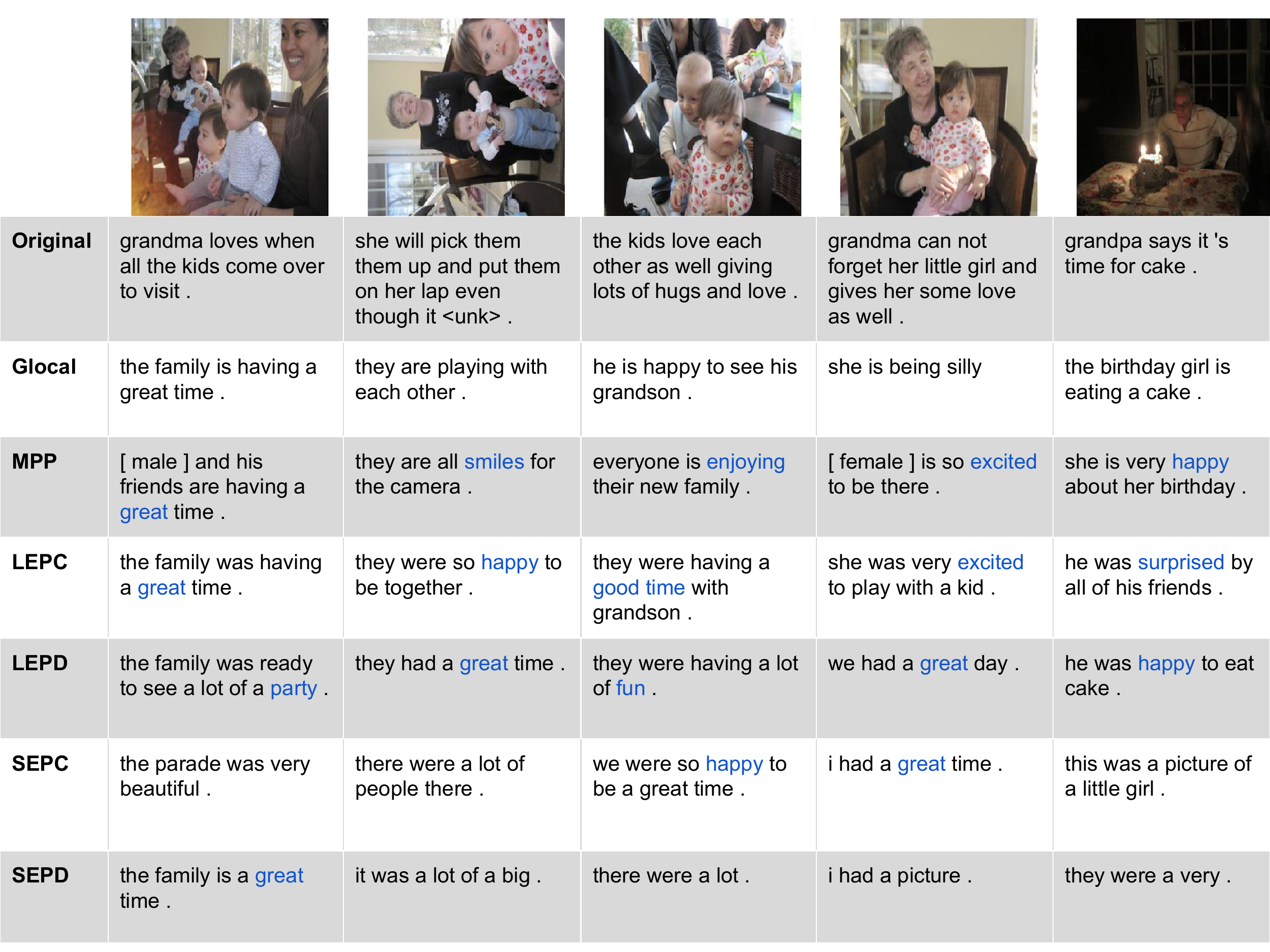}
\caption{ Comparison of generated \textit{stories} from all the described models. }
\label{fig:analysis}
\end{figure*}

\section{Experiments and Results}

This section presents the experimental setup for the models described in Section \ref{sec:models}. Each of the models are incremental extensions over the baseline glocal model. The hyperparameters used for this are as follows.

\paragraph{Hyperparameters: } The hidden size of the Bi-LSTM encoder of the story to capture context is 1024.
The dimensionality of the glocal context vector $\boldsymbol{z}_k$ is 2048.
A dropout layer of 50\% is applied post the fully connected layer to obtain the image features and after the global features obtained from Bi-LSTM which is 2 layered.
The word embedding dimension used is 256.
The learning rate  is 1e-3 with a weight decay of 1e-5. Adam optimizer is used with batch normalization and a momentum of 0.01.
Weighting the loss functions differently is done to penalize the model more if the decoding is at fault as compared to not predicting the personality of the story. $\alpha$ is set to 0.5 and each of the individual personality losses are weighted by a factor of 0.1.

The rest of the 5 models use the same hyperparameter setting with an exception to word embedding dimension.
The average personality ($\boldsymbol{\mathcal{P}}$) and the average story ($\boldsymbol{\mathcal{S}}$) representations are obtained from pre-trained BERT model.
Hence this is a 768 dimensional vector.
In order to perform the stripping of the story feature and adding the personality features to the word embeddings in the decoder, the word embedding dimension is matched to 768 in the SEPD model.

\begin{table}[h!]
\centering
\begin{tabular}{l | r | r | r | r | r }
\hline
Model & C1 & C2 & C3 & C4 & C5 \\
\hline
\textbf{Glocal} & 69.90 & 73.29 & 51.55 & 34.91 & 65.86 \\
\textbf{MPP} & 69.35 & 72.44 & 47.54 & 33.83 & 58.49 \\
\textbf{LEPC} & 70.10 & 73.24 & 52.13 & 34.59 & 66.42 \\
\textbf{LEPD} & 76.44 & 79.20 & 33.71 & 34.02 & 67.13 \\
\textbf{SEPC} & 76.76 & 77.00 & 32.84 & 44.53 & 60.08 \\
\textbf{SEPD} & 78.14 & 79.44 & 31.33 & 34.99 & 73.88 \\
\hline
\end{tabular}
\caption{Performance (in terms of accuracy) of generated stories to capture persona}
\label{tab:res}
\end{table}

\begin{table}[h]
\centering
\begin{tabular}{l | r }
\hline
Model & ROUGE\_L \\
\hline
Glocal &  0.1805\\
MPP &  0.1713\\
LEPC &  0.1814\\
LEPD &  0.1731\\
SEPC &  0.1665\\
SEPD &  0.1689\\
\hline
\end{tabular}
\caption{ROUGE\_L scores for the generated stories by each of our models}
\label{tab:meteor}
\end{table}

\subsection{Quantitative Results}
We perform two sets of experiments: (1) evaluating the performance of the models on capturing the personalities in the story and 
(2) performance of story generation.
The former evaluation is performed using the pre-trained classifiers (\ref{sec:classification}) on the personality dataset.
We calculate the classification accuracy of the generated stories of the test set for the desired target personality. However, we need to note that the classification error of the models trained is reflected in this result as well.
This evaluation is done at a sentence level i.e accuracy is calculated over each sentence of the story (each sentence of the story has the same target personality as that of the entire story).
The performance of the generation is evaluated using the ROUGE score \footnote{We use the implementation from \url{https://github.com/Maluuba/nlg-eval}}. Although this captures the generic aspect of generation, the metric explicitly does not evaluate whether the story is generated on a conditioned personality. In future, we would also like to look at automatic evaluation of the generated stories with respect to incorporation of personalities.

Table \ref{tab:res} shows the results of classification accuracy for each of the five personalities.
Table \ref{tab:meteor} shows the results of ROUGE\_L evaluation.
We acknowledge that there would be a deviation to this automatic score since optimizing the gold standard generation of story from training data is not our end goal.
Rather our models make use of two distinct datasets and learn to transfer the traits annotated in personality dialog dataset into the visual story telling dataset.

Despite this, we notice that LEPC model gives comparative results to that of the glocal model in terms of story generation.
It is noticed that LEPC model also gives slight improvement on the classification accuracies for most of the clusters (each cluster representing a personality).
However this is an insufficient result to generalize that incorporating personality at context level performs better than that at the word level since the inverted stance is observed in SEPC and SEPD models.
We plan to investigate this further by performing ablations and examine which operation is causing these models to perform weakly.
Note that the SEPC model performs the best in incorporating personality in three of the five personality types.
But this model takes a hit in the automatic score.
This is because our generative models are dealing with competing losses or reconstruction of classification.

\subsection{Qualitative Results}

We present an example of the story generated by each of the models proposed in Figure \ref{fig:analysis}. This example belongs to persona in cluster \textbf{C3}. The words corresponding to this cluster are highlighted with blue color in the persona conditioned generation of the stories. The main observation is that all of the five sentences in the story contain a word relevant to \textit{happiness} for each of the MPP, LEPC and LEPD models. SEPC and SEPD models capture these happiness features in only two and one sentences respectively. The glocal model does not cater explicitly to the personality while our proposed models attempt to capture the persona tone in generation. This is observed in the fourth generated sentence in the sequence by each of our proposed models. While the glocal model uses the word \textit{`silly'}, our models capture the tone and generate \textit{`excited'} and \textit{`great'}. Similarly for the fifth sentence, MPP, LEPC and LEPD generate \textit{`happy'}, \textit{`surprised'} and \textit{`happy'} respectively.

It is observed that in most generated stories, the language model has taken a rough hit in the SEPD model. This is also substantiated in Figure \ref{fig:analysis}. This seems to be due to stripping away the essential word embedding features that contribute to linguistic priors or language model. This could be potentially corrected by retaining the word embedding feature as is and augmenting it with the stripped features. Having presented these results, we notice that there is a significant scope for improving the generation of the story while capturing high level persona traits in generation.

\section{Conclusions and Future Work}

Automatic storytelling is a creative writing task that has long been the dream of text generation models.
The voice conveying this story is the narrative style and this can be attributed to different personalities, moods, situations etc.
In the case of persona based visual storytelling, this voice not only is aware of the grounded content to be conveyed in the images, but also has a model to steer the words in the narrative to characterize the persona.

A key challenge here is that there is no targeted data for this specific task.
Hence we leverage annotations of persona from an external persona based dialog dataset and apply it on the visual storytelling dataset.
We address this task of attribution of a personality while generating a grounded story by simple techniques of incorporating persona information in our encoder-decoder architecture.
We propose five simple incremental extensions to the baseline model that captures the personality.
Quantitatively, our results show that the LEPC model is improving upon the accuracy while at the same time not dropping the automatic scores.
We also observe that the persona induced models are generating at least one word per sentence in the story that belong to that particular persona.
While automatically evaluating this can be tricky, we adapt a classification based evaluation of whether the generated output belongs to the persona class or not.
In the future, we hope to also perform human evaluations for measuring both the target personality type of the generated and story and its coherence.

There is yet a lot of scope in incorporating the persona in the word embeddings. This is an ongoing work and we plan on investigating the relatively poor ROUGE performance of the SEPC and SEPD models and rectify them by equipping them with language model information. We also plan to work towards a stable evaluation protocol for this task in the future.


\bibliography{acl2019}

\begin{thebibliography}{39}
\expandafter\ifx\csname natexlab\endcsname\relax\def\natexlab#1{#1}\fi

\bibitem[{Cavazza et~al.(2009)Cavazza, Pizzi, Charles, Vogt, and
  Andr{\'e}}]{cavazza2009emotional}
Marc Cavazza, David Pizzi, Fred Charles, Thurid Vogt, and Elisabeth Andr{\'e}.
  2009.
\newblock Emotional input for character-based interactive storytelling.
\newblock In \emph{Proceedings of The 8th International Conference on
  Autonomous Agents and Multiagent Systems-Volume 1}, pages 313--320.
  International Foundation for Autonomous Agents and Multiagent Systems.

\bibitem[{Chandrasekaran et~al.(2018)Chandrasekaran, Parikh, and
  Bansal}]{chandrasekaran2018punny}
Arjun Chandrasekaran, Devi Parikh, and Mohit Bansal. 2018.
\newblock Punny captions: Witty wordplay in image descriptions.
\newblock In \emph{Proceedings of the 2018 Conference of the North American
  Chapter of the Association for Computational Linguistics: Human Language
  Technologies, Volume 2 (Short Papers)}, pages 770--775.

\bibitem[{Chandu et~al.(2019)Chandu, Black, and Nyberg}]{chandustoryboarding}
Khyathi Chandu, Alan~W Black, and Eric Nyberg. 2019.
\newblock Storyboarding of recipes: Grounded contextual generation.
\newblock In \emph{Proceedings of the 57th Annual Meeting of the Association
  for Computational Linguistics}, Florence, Italy. Association for
  Computational Linguistics.

\bibitem[{Das et~al.(2017)Das, Kottur, Gupta, Singh, Yadav, Moura, Parikh, and
  Batra}]{das2017visual}
Abhishek Das, Satwik Kottur, Khushi Gupta, Avi Singh, Deshraj Yadav,
  Jos{\'e}~MF Moura, Devi Parikh, and Dhruv Batra. 2017.
\newblock Visual dialog.
\newblock In \emph{Proceedings of the IEEE Conference on Computer Vision and
  Pattern Recognition}, volume~2.

\bibitem[{Devlin et~al.(2018)Devlin, Chang, Lee, and
  Toutanova}]{devlin2018bert}
Jacob Devlin, Ming-Wei Chang, Kenton Lee, and Kristina Toutanova. 2018.
\newblock Bert: Pre-training of deep bidirectional transformers for language
  understanding.
\newblock \emph{arXiv preprint arXiv:1810.04805}.

\bibitem[{Fan et~al.(2018)Fan, Lewis, and Dauphin}]{fan2018hierarchical}
Angela Fan, Mike Lewis, and Yann Dauphin. 2018.
\newblock Hierarchical neural story generation.
\newblock In \emph{Proceedings of the 56th Annual Meeting of the Association
  for Computational Linguistics (Volume 1: Long Papers)}, pages 889--898.

\bibitem[{Fu et~al.(2018)Fu, Tan, Peng, Zhao, and Yan}]{fu2018style}
Zhenxin Fu, Xiaoye Tan, Nanyun Peng, Dongyan Zhao, and Rui Yan. 2018.
\newblock Style transfer in text: Exploration and evaluation.
\newblock In \emph{Thirty-Second AAAI Conference on Artificial Intelligence}.

\bibitem[{Gella et~al.(2018)Gella, Lewis, and Rohrbach}]{gella2018dataset}
Spandana Gella, Mike Lewis, and Marcus Rohrbach. 2018.
\newblock A dataset for telling the stories of social media videos.
\newblock In \emph{Proceedings of the 2018 Conference on Empirical Methods in
  Natural Language Processing}, pages 968--974.

\bibitem[{Gonzalez-Rico and Fuentes-Pineda(2018)}]{gonzalez2018contextualize}
Diana Gonzalez-Rico and Gibran Fuentes-Pineda. 2018.
\newblock Contextualize, show and tell: a neural visual storyteller.
\newblock \emph{arXiv preprint arXiv:1806.00738}.

\bibitem[{Guan et~al.(2018)Guan, Wang, and Huang}]{guan2018story}
Jian Guan, Yansen Wang, and Minlie Huang. 2018.
\newblock Story ending generation with incremental encoding and commonsense
  knowledge.
\newblock \emph{arXiv preprint arXiv:1808.10113}.

\bibitem[{He et~al.(2016)He, Zhang, Ren, and Sun}]{he2016deep}
Kaiming He, Xiangyu Zhang, Shaoqing Ren, and Jian Sun. 2016.
\newblock Deep residual learning for image recognition.
\newblock In \emph{Proceedings of the IEEE conference on computer vision and
  pattern recognition}, pages 770--778.

\bibitem[{Hossain et~al.(2019)Hossain, Sohel, Shiratuddin, and
  Laga}]{hossain2019comprehensive}
MD~Hossain, Ferdous Sohel, Mohd~Fairuz Shiratuddin, and Hamid Laga. 2019.
\newblock A comprehensive survey of deep learning for image captioning.
\newblock \emph{ACM Computing Surveys (CSUR)}, 51(6):118.

\bibitem[{Hsu et~al.(2018)Hsu, Chen, Hsieh, and Ku}]{hsu2018using}
Chao-Chun Hsu, Szu-Min Chen, Ming-Hsun Hsieh, and Lun-Wei Ku. 2018.
\newblock Using inter-sentence diverse beam search to reduce redundancy in
  visual storytelling.
\newblock \emph{arXiv preprint arXiv:1805.11867}.

\bibitem[{Hu et~al.(2017)Hu, Yang, Liang, Salakhutdinov, and
  Xing}]{hu2017toward}
Zhiting Hu, Zichao Yang, Xiaodan Liang, Ruslan Salakhutdinov, and Eric~P Xing.
  2017.
\newblock Toward controlled generation of text.
\newblock In \emph{Proceedings of the 34th International Conference on Machine
  Learning-Volume 70}, pages 1587--1596. JMLR. org.

\bibitem[{Huang et~al.(2016)Huang, Ferraro, Mostafazadeh, Misra, Agrawal,
  Devlin, Girshick, He, Kohli, Batra et~al.}]{huang2016visual}
Ting-Hao~Kenneth Huang, Francis Ferraro, Nasrin Mostafazadeh, Ishan Misra,
  Aishwarya Agrawal, Jacob Devlin, Ross Girshick, Xiaodong He, Pushmeet Kohli,
  Dhruv Batra, et~al. 2016.
\newblock Visual storytelling.
\newblock In \emph{Proceedings of the 2016 Conference of the North American
  Chapter of the Association for Computational Linguistics: Human Language
  Technologies}, pages 1233--1239.

\bibitem[{Kim et~al.(2018)Kim, Heo, Son, Park, and Zhang}]{kim2018glac}
Taehyeong Kim, Min-Oh Heo, Seonil Son, Kyoung-Wha Park, and Byoung-Tak Zhang.
  2018.
\newblock Glac net: Glocal attention cascading networks for multi-image cued
  story generation.
\newblock \emph{arXiv preprint arXiv:1805.10973}.

\bibitem[{Lample et~al.(2019)Lample, Subramanian, Smith, Denoyer, Ranzato, and
  Boureau}]{lample2018multipleattribute}
Guillaume Lample, Sandeep Subramanian, Eric Smith, Ludovic Denoyer,
  Marc'Aurelio Ranzato, and Y-Lan Boureau. 2019.
\newblock \href {https://openreview.net/forum?id=H1g2NhC5KQ}
  {Multiple-attribute text rewriting}.
\newblock In \emph{International Conference on Learning Representations}.

\bibitem[{Lau and Baldwin(2016)}]{lau2016empirical}
Jey~Han Lau and Timothy Baldwin. 2016.
\newblock An empirical evaluation of doc2vec with practical insights into
  document embedding generation.
\newblock \emph{ACL 2016}, page~78.

\bibitem[{Li et~al.(2016)Li, Galley, Brockett, Spithourakis, Gao, and
  Dolan}]{li2016persona}
Jiwei Li, Michel Galley, Chris Brockett, Georgios Spithourakis, Jianfeng Gao,
  and Bill Dolan. 2016.
\newblock A persona-based neural conversation model.
\newblock In \emph{Proceedings of the 54th Annual Meeting of the Association
  for Computational Linguistics (Volume 1: Long Papers)}, volume~1, pages
  994--1003.

\bibitem[{Li et~al.(2018)Li, Jia, He, and Liang}]{li2018delete}
Juncen Li, Robin Jia, He~He, and Percy Liang. 2018.
\newblock Delete, retrieve, generate: a simple approach to sentiment and style
  transfer.
\newblock In \emph{Proceedings of the 2018 Conference of the North American
  Chapter of the Association for Computational Linguistics: Human Language
  Technologies, Volume 1 (Long Papers)}, pages 1865--1874.

\bibitem[{Lin(2004)}]{lin2004rouge}
Chin-Yew Lin. 2004.
\newblock Rouge: A package for automatic evaluation of summaries.
\newblock \emph{Text Summarization Branches Out}.

\bibitem[{Logeswaran et~al.(2018)Logeswaran, Lee, and
  Bengio}]{logeswaran2018content}
Lajanugen Logeswaran, Honglak Lee, and Samy Bengio. 2018.
\newblock Content preserving text generation with attribute controls.
\newblock In \emph{Advances in Neural Information Processing Systems}, pages
  5103--5113.

\bibitem[{Lukin et~al.(2018)Lukin, Hobbs, and Voss}]{lukin2018pipeline}
Stephanie Lukin, Reginald Hobbs, and Clare Voss. 2018.
\newblock A pipeline for creative visual storytelling.
\newblock In \emph{Proceedings of the First Workshop on Storytelling}, pages
  20--32.

\bibitem[{Martin et~al.(2018)Martin, Ammanabrolu, Wang, Hancock, Singh,
  Harrison, and Riedl}]{martin2018event}
Lara~J Martin, Prithviraj Ammanabrolu, Xinyu Wang, William Hancock, Shruti
  Singh, Brent Harrison, and Mark~O Riedl. 2018.
\newblock Event representations for automated story generation with deep neural
  nets.
\newblock In \emph{Thirty-Second AAAI Conference on Artificial Intelligence}.

\bibitem[{Mostafazadeh et~al.(2017)Mostafazadeh, Brockett, Dolan, Galley, Gao,
  Spithourakis, and Vanderwende}]{mostafazadeh2017image}
Nasrin Mostafazadeh, Chris Brockett, Bill Dolan, Michel Galley, Jianfeng Gao,
  Georgios~P Spithourakis, and Lucy Vanderwende. 2017.
\newblock Image-grounded conversations: Multimodal context for natural question
  and response generation.
\newblock \emph{arXiv preprint arXiv:1701.08251}.

\bibitem[{Mostafazadeh et~al.(2016)Mostafazadeh, Grealish, Chambers, Allen, and
  Vanderwende}]{mostafazadeh2016caters}
Nasrin Mostafazadeh, Alyson Grealish, Nathanael Chambers, James Allen, and Lucy
  Vanderwende. 2016.
\newblock Caters: Causal and temporal relation scheme for semantic annotation
  of event structures.
\newblock In \emph{Proceedings of the Fourth Workshop on Events}, pages 51--61.

\bibitem[{Peng et~al.(2018)Peng, Ghazvininejad, May, and
  Knight}]{peng2018towards}
Nanyun Peng, Marjan Ghazvininejad, Jonathan May, and Kevin Knight. 2018.
\newblock Towards controllable story generation.
\newblock In \emph{Proceedings of the First Workshop on Storytelling}, pages
  43--49.

\bibitem[{Prabhumoye et~al.(2018)Prabhumoye, Tsvetkov, Salakhutdinov, and
  Black}]{prabhumoye2018style}
Shrimai Prabhumoye, Yulia Tsvetkov, Ruslan Salakhutdinov, and Alan~W Black.
  2018.
\newblock Style transfer through back-translation.
\newblock In \emph{Proceedings of the 56th Annual Meeting of the Association
  for Computational Linguistics (Volume 1: Long Papers)}, pages 866--876.

\bibitem[{Rao and Tetreault(2018)}]{rao2018dear}
Sudha Rao and Joel Tetreault. 2018.
\newblock Dear sir or madam, may i introduce the gyafc dataset: Corpus,
  benchmarks and metrics for formality style transfer.
\newblock In \emph{Proceedings of the 2018 Conference of the North American
  Chapter of the Association for Computational Linguistics: Human Language
  Technologies, Volume 1 (Long Papers)}, pages 129--140.

\bibitem[{Rishes et~al.(2013)Rishes, Lukin, Elson, and
  Walker}]{rishes2013generating}
Elena Rishes, Stephanie~M Lukin, David~K Elson, and Marilyn~A Walker. 2013.
\newblock Generating different story tellings from semantic representations of
  narrative.
\newblock In \emph{International Conference on Interactive Digital
  Storytelling}, pages 192--204. Springer.

\bibitem[{Sharma et~al.(2018)Sharma, Allen, Bakhshandeh, and
  Mostafazadeh}]{sharma2018tackling}
Rishi Sharma, James Allen, Omid Bakhshandeh, and Nasrin Mostafazadeh. 2018.
\newblock Tackling the story ending biases in the story cloze test.
\newblock In \emph{Proceedings of the 56th Annual Meeting of the Association
  for Computational Linguistics (Volume 2: Short Papers)}, volume~2, pages
  752--757.

\bibitem[{Shen et~al.(2017)Shen, Lei, Barzilay, and Jaakkola}]{shen2017style}
Tianxiao Shen, Tao Lei, Regina Barzilay, and Tommi Jaakkola. 2017.
\newblock Style transfer from non-parallel text by cross-alignment.
\newblock In \emph{Advances in neural information processing systems}, pages
  6830--6841.

\bibitem[{Shuster et~al.(2018)Shuster, Humeau, Bordes, and
  Weston}]{shuster2018engaging}
Kurt Shuster, Samuel Humeau, Antoine Bordes, and Jason Weston. 2018.
\newblock Engaging image chat: Modeling personality in grounded dialogue.
\newblock \emph{arXiv preprint arXiv:1811.00945}.

\bibitem[{Smilevski et~al.(2018)Smilevski, Lalkovski, and
  Madzarov}]{smilevski2018stories}
Marko Smilevski, Ilija Lalkovski, and Gjorgi Madzarov. 2018.
\newblock Stories for images-in-sequence by using visual and narrative
  components.
\newblock \emph{arXiv preprint arXiv:1805.05622}.

\bibitem[{Wang and Wen(2015)}]{wang2015can}
William~Yang Wang and Miaomiao Wen. 2015.
\newblock I can has cheezburger? a nonparanormal approach to combining textual
  and visual information for predicting and generating popular meme
  descriptions.
\newblock In \emph{Proceedings of the 2015 Conference of the North American
  Chapter of the Association for Computational Linguistics: Human Language
  Technologies}, pages 355--365.

\bibitem[{Wiseman et~al.(2017)Wiseman, Shieber, and
  Rush}]{wiseman2017challenges}
Sam Wiseman, Stuart Shieber, and Alexander Rush. 2017.
\newblock Challenges in data-to-document generation.
\newblock In \emph{Proceedings of the 2017 Conference on Empirical Methods in
  Natural Language Processing}, pages 2253--2263.

\bibitem[{Wu et~al.(2017)Wu, Teney, Wang, Shen, Dick, and van~den
  Hengel}]{wu2017visual}
Qi~Wu, Damien Teney, Peng Wang, Chunhua Shen, Anthony Dick, and Anton van~den
  Hengel. 2017.
\newblock Visual question answering: A survey of methods and datasets.
\newblock \emph{Computer Vision and Image Understanding}, 163:21--40.

\bibitem[{Zhang et~al.(2018)Zhang, Dinan, Urbanek, Szlam, Kiela, and
  Weston}]{zhang2018personalizing}
Saizheng Zhang, Emily Dinan, Jack Urbanek, Arthur Szlam, Douwe Kiela, and Jason
  Weston. 2018.
\newblock Personalizing dialogue agents: I have a dog, do you have pets too?
\newblock \emph{arXiv preprint arXiv:1801.07243}.

\bibitem[{Zhou et~al.(2018)Zhou, Huang, Zhang, Zhu, and
  Liu}]{zhou2018emotional}
Hao Zhou, Minlie Huang, Tianyang Zhang, Xiaoyan Zhu, and Bing Liu. 2018.
\newblock Emotional chatting machine: Emotional conversation generation with
  internal and external memory.
\newblock In \emph{Thirty-Second AAAI Conference on Artificial Intelligence}.

\end{thebibliography}
\bibliographystyle{acl_natbib}

\appendix



\end{document}